\documentclass{article}

\usepackage[final]{nips_2017}

\usepackage[utf8]{inputenc} 
\usepackage[T1]{fontenc}    
\usepackage{hyperref}       
\usepackage{url}            
\usepackage{booktabs}       
\usepackage{amsfonts}       
\usepackage{nicefrac}       
\usepackage{microtype}      
\usepackage{graphicx}
\usepackage{subcaption}
\newcommand*\samethanks[1][\value{footnote}]{\footnotemark[#1]}
\title{Deep Fault Analysis and Subset Selection in Solar Power Grids}
\author{
Biswarup Bhattacharya \samethanks\\
University of Southern California\\
Los Angeles, CA 90089. USA.\\
Email: bbhattac@usc.edu
\And
Abhishek Sinha \samethanks\\
Adobe Systems Incorporated\\
Noida, UP 201301. India.\\
Email: abhishek.sinha94@gmail.com
}

\begin{document}
\maketitle
\begin{abstract}
Non-availability of reliable and sustainable electric power is a major problem in the developing world. Renewable energy sources like solar are not very lucrative in the current stage due to various uncertainties like weather, storage, land use among others. There also exists various other issues like mis-commitment of power, absence of intelligent fault analysis, congestion, etc. In this paper, we propose a novel deep learning-based system for predicting faults and selecting power generators optimally so as to reduce costs and ensure higher reliability in solar power systems. The results are highly encouraging and they suggest that the approaches proposed in this paper have the potential to be applied successfully in the developing world.
\end{abstract}

\section{Introduction}
Electric power grids are an essential component of modern society. A reliable power supply is conducive to development of infrastructure and improving people's quality of life. The demand for power has increased to the point that resources need to be conserved carefully to generate power in optimal amounts so that no power is wasted and there is no shortage. With the help of technology like synchrophasors, the magnitude and angle of each phase of the three phase voltage and/or current, frequency, rate of change of frequency and angular separation at every few millisecond interval (say 40 milliseconds) can be monitored \cite{synchrophasor, pmu}. This data can be leveraged to incorporate intelligence into the system to make better predictions and design better schemes for generation and fault analysis.

In developing countries like India, renewable energy is often wasted due to mismanagement and due to irregularities in the generation process \cite{renewindia}. For example, in case of solar energy, storage of generated power becomes an issue and non-optimal choices lead to wastage of generated power. Power outages are common in developing countries due to faults appearing in the system which could not be predicted earlier. Often the load forecast does not match the actual scenario and that leads to problems like congestion and faults. 

In this paper, we have proposed a method to analyze faults using LSTMs which can analyze the grid information at any given time and determine the health of the grid; and we have also proposed a method to simulate various conditions including stimuli like generator supply, weather and load demand using Siemens PSS/E software to determine the optimal number of solar energy generator subsets to be selected for sustainable power generation. We designed a grid ourselves to simulate conditions and run experiments.

\section{Fault Analysis}
In the current situation of Indian electrical power grids, when a small disturbance is seen at a dispatch center, then the protocol is to generate a report and check with other dispatch centers. If the disturbance is found to be not local, then further diagnostics are conducted to find the source of the problem and repair it. In our approach, we monitor the grid continuously and then in the case of a fault determine the type of the fault as and when it occurs. We additionally also find out the exact bus line where the fault has occurred which is sure to aid in rapid recovery and maintenance. This automated process lessens human supervision and enables better management of the grid.

\subsection{Data}
We generated data for this problem using the Siemens PSS/E software \cite{siemens} and PowerWorld Simulator \cite{powersim}. The Siemens PSS/E software package can do fast and robust power flow solution for network models up to 200,000 buses. It has an array of useful features for dynamic and transient analysis. It also has a useful scripting system called {\tt{psspy}} which we utilized for running the experiments. The detailed process of data collection has been described in \cite{bhattacharya1}.

\subsection{Determining the fault type}
We analyzed four types of faults: $3\phi$ bus fault, branch trip fault, LL fault, and LG fault. The first two are symmetrical faults, whereas the last two are unsymmetrical faults. We designed a grid of 23 buses and we simulated these 4 faults to generate ground truth data for our machine learning system. We simulated every type of fault when the fault occurs at each individual bus. For each bus, the simulation was run 100 times to generate different variations so that there are enough examples.

The model consists of 100 unfoldings in time of LSTM cells corresponding to 100 time varying voltage values for each of the bus. The input to each LSTM cell is a vector of bus voltages at every time step. Both the hidden vector and output vectors are of size 128. The output vector of the last cell contains the temporal information present in the entire data. With this information, we can classify the type of fault. For the classification, a second model is built entirely of fully connected layers. The model consists of 1 hidden layer consisting of 64 neurons followed by an output layer of size 2. The output obtained determines whether the fault belongs to LL or LG class.

With LSTM the classification accuracy is around 95\%, an improvement of around 6\% from the baseline SVM model. The plot of accuracy is shown in Figure 1(a).

\subsection{Deducing the faulted bus line number}
Once the type of fault was determined by the classification model, different models were constructed for each of the different fault types to determine the faulted bus line number. A similar LSTM network was used with the same input of the bus voltages. The classifier was trained to output the bus number corresponding to where the fault had been triggered or an output of $0$ if the network data corresponded to a non-faulty one where no fault had been triggered on any bus line.

The classifier returned an accuracy of 97\% corresponding to predicting the bus number. The bus number was outputted or classified as ``$0$'' in case of non-faulty data. The accuracy plot has been shown in Figure 1(b).

\begin{figure}[h]
\begin{subfigure}{0.45\textwidth}
\includegraphics[width=\linewidth, keepaspectratio]{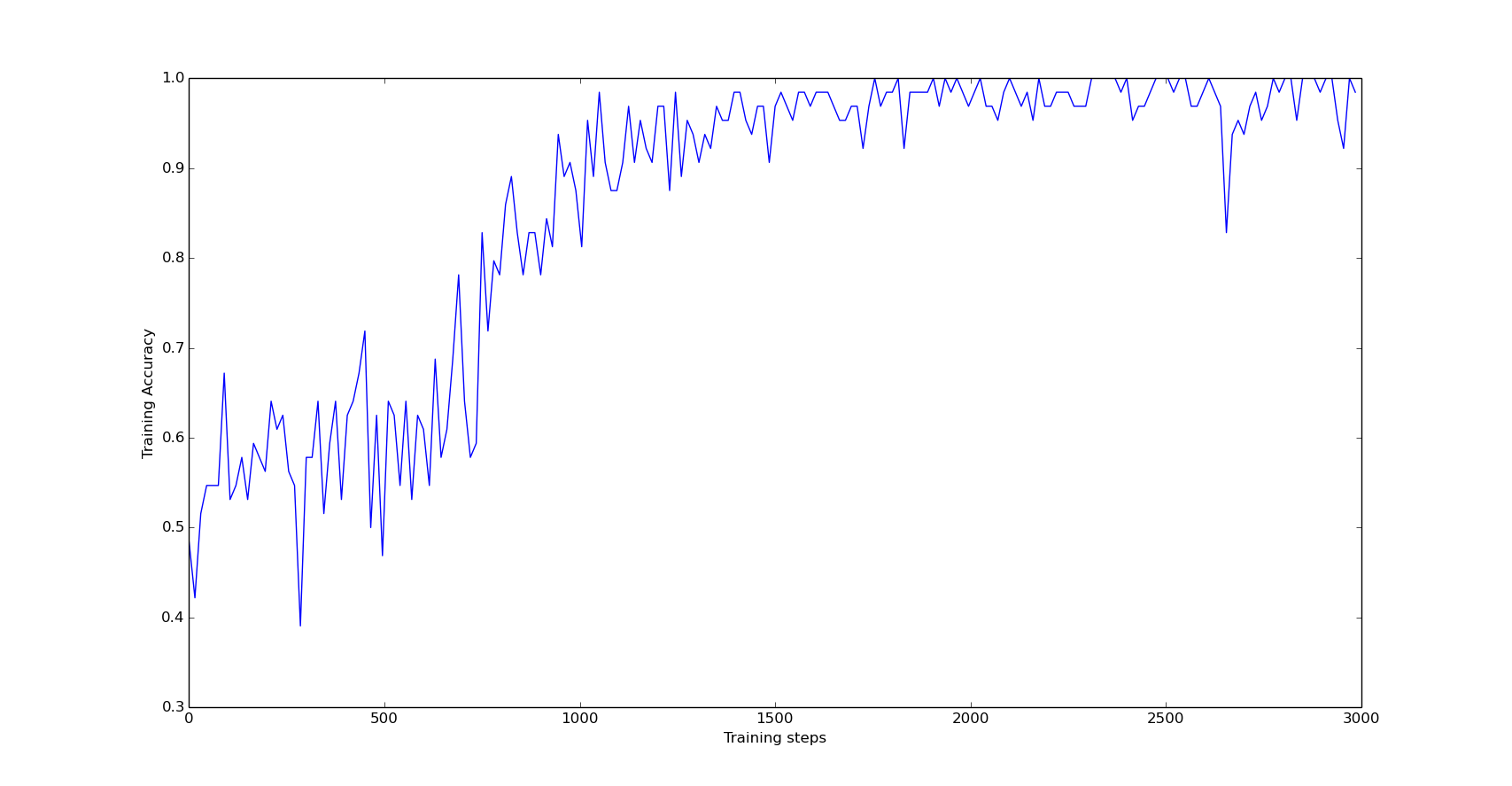}
\caption{LSTM Accuracy for Fault Classification}
\end{subfigure}
\hfill
\begin{subfigure}{0.45\textwidth}
\includegraphics[width=\linewidth, keepaspectratio]{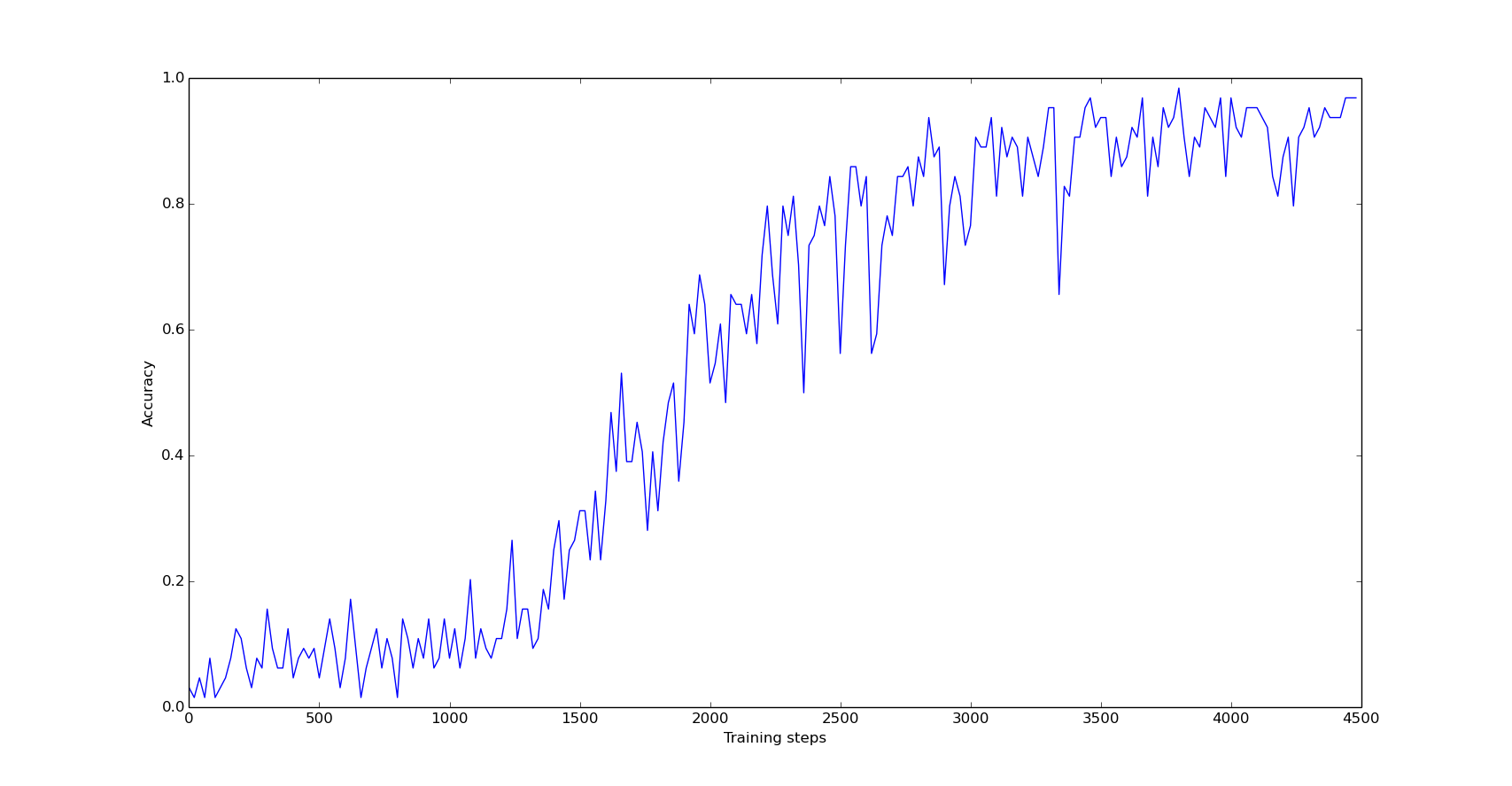}
\caption{Accuracy for Bus Line Classification}
\end{subfigure}
\caption{Accuracy Plots for Fault Analysis}
\end{figure}

\section{Subset Selection}
In order to participate in the electricity grids markets with a day-ahead commitment \cite{al} similar to the usual case, solar power generators have to guarantee a power supply commitment in competition with the generally more reliable traditional (fossil-fuel based) generators. Generally, in case of an imbalance, renewable generators have to pay for the difference in power contract and the actual power output. We propose a technique that can reduce this power forecasting uncertainty for renewables while keeping in mind the factors of congestion and load demand. Our task is to ensure economic dispatch.

\subsection{Data}
Due to lack of availability of free to use, well formatted, desired data, we collected data for 16 days from March 1, 2017 to March 14, 2017 from the IIT Kharagpur Electrical Engineering department rooftop solar panels.

\subsection{Congestion prediction}
Transmission congestion occurs when there is insufficient energy to meet the demands of all customers. This problem is especially prevalent in developing countries. We built two predictive models - one for when the actual generated solar power is known, and one for when it is not known. In reality, we know just the power committed by the solar power generators and not the actual power generated in the future. Hence, using weather information, this power was predicted. Finally, the congestion prediction was done by using a neural network with 1 hidden layer. The test accuracy was around 97-98\% for the known case, and around 91\% for the unknown solar power case. In both cases, the neural network model achieved an accuracy improvement of around 5\% over the baseline SVM model.

\subsubsection{Data Generation for actual subset selection}
The main motivation behind the data generation process was to simulate the various scenarios which can happen due to only a subset of the generators being ``on''. Our objective was to find the subset of the generators to be switched ``on'' to meet the demand at the least cost. The whole set of generators included both solar as well as non-conventional power generators. Initially all solar generators were kept switched ``on'' and then a load flow was run to get the state at $t=0^{+}$. Then, at $t=1$, a subset of the generators was switched off and the load state was run again. We were trying to observe the one step jumps and its effect on the grid. We also simultaneously checked for congestion. The process is explained in detail in \cite{bhattacharya2}.

\subsection{Choosing the optimal subset}
The task that we are trying to solve in this part is the selection of the optimal subset of solar generators that minimize the total penalty occurred in using the solar generators. We considered two types of penalty - power mis-commitment penalty and congestion penalty.

\subsubsection{Power mis-commitment penalty}
Uncertainty in weather implies erratic solar power generation which leads to the power generators often being unable to generate the amount of power which they had promised to generate. The difference in the promised and the delivered power is covered by the non-renewable source generators but it leads to an extra cost incurred by the grid operators. The larger the difference is between the committed and the delivered powers, the greater is the penalty incurred. Thus to choose an optimal subset one of our aims should be to choose the subset that shows lesser variance in the prediction depending on the current weather conditions. Thus the loss $L_1$ used to train the network is given by:
\begin{equation} 
L_1 = (\textnormal{predicted power} - \textnormal{actual power generated})
\end{equation}

\subsubsection{Penalty incurred due to congestion in the network}
In certain cases, even if the mis-prediction penalty for choosing a subset is lower than choosing another subset, choosing the other subset might cause a congestion in the network which is extremely undesirable. Hence, we incorporated a second penalty. The congestion loss $L_2$ is calculated as follows: if a subset causes a congestion in the network then $L_2 = 50$ otherwise $L_2 = 0$.

\subsection{The Model}
For all the possible subsets of generators, the $L_1$ and $L_2$ values were computed and scaled. The input to the model are the bus voltages before any subset is chosen, and a vector of size corresponding to the number of solar generators with the elements of the vector being 1 or 0 depending on whether the generator forms a part of the subset. The total loss corresponding to the subset is $L_1+L_2$.

The model would be expected to output the total loss incurred in selecting the input subset. Thus the loss function ($\mathcal{L}$) that the model tries to minimize is given by:
\begin{equation}
\mathcal{L} = (\textnormal{model output} - (L_1+L_2))^2
\end{equation}

The model consisted of $1$ hidden layer with $200$ hidden neurons. Adam optimizer \cite{adam} was used to train the model and minimize the loss $\mathcal{L}$.

\subsection{Results}
After training the model for around $2500$ steps, the training minimum $L_2$ loss = $35$, the training minimum $L_1$ loss = $6$, the test $L_2$ loss = $59$, and the test $L_1$ loss = $8$. The plots for the training and test $L_2$ losses have been shown in Figure 2.

\begin{figure}[h]
\begin{subfigure}{0.45\textwidth}
\includegraphics[width=\linewidth, keepaspectratio]{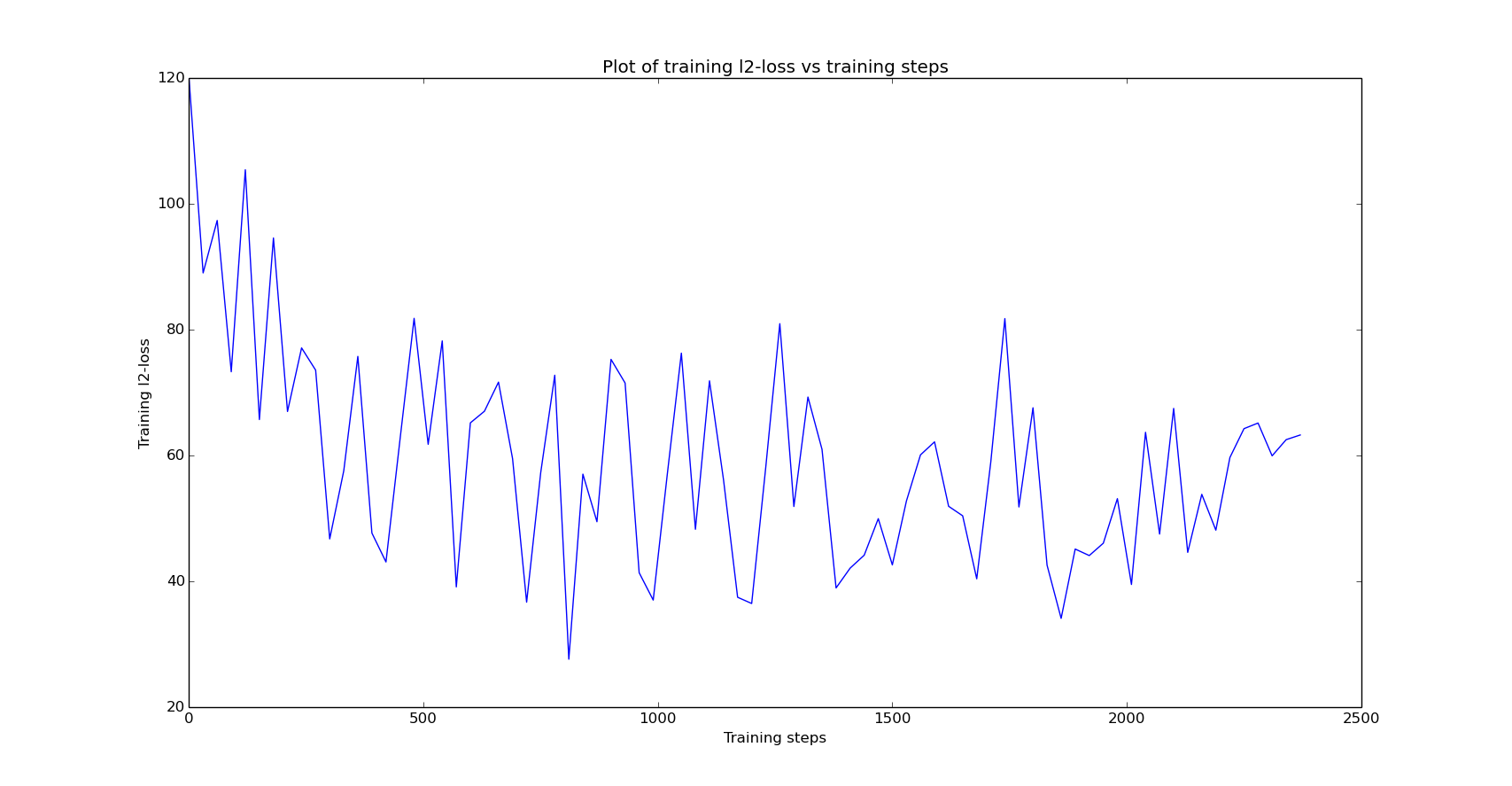}
\caption{Plot of training $L_2$ loss with progress in training}
\end{subfigure}
\hfill
\begin{subfigure}{0.45\textwidth}
\includegraphics[width=\linewidth, keepaspectratio]{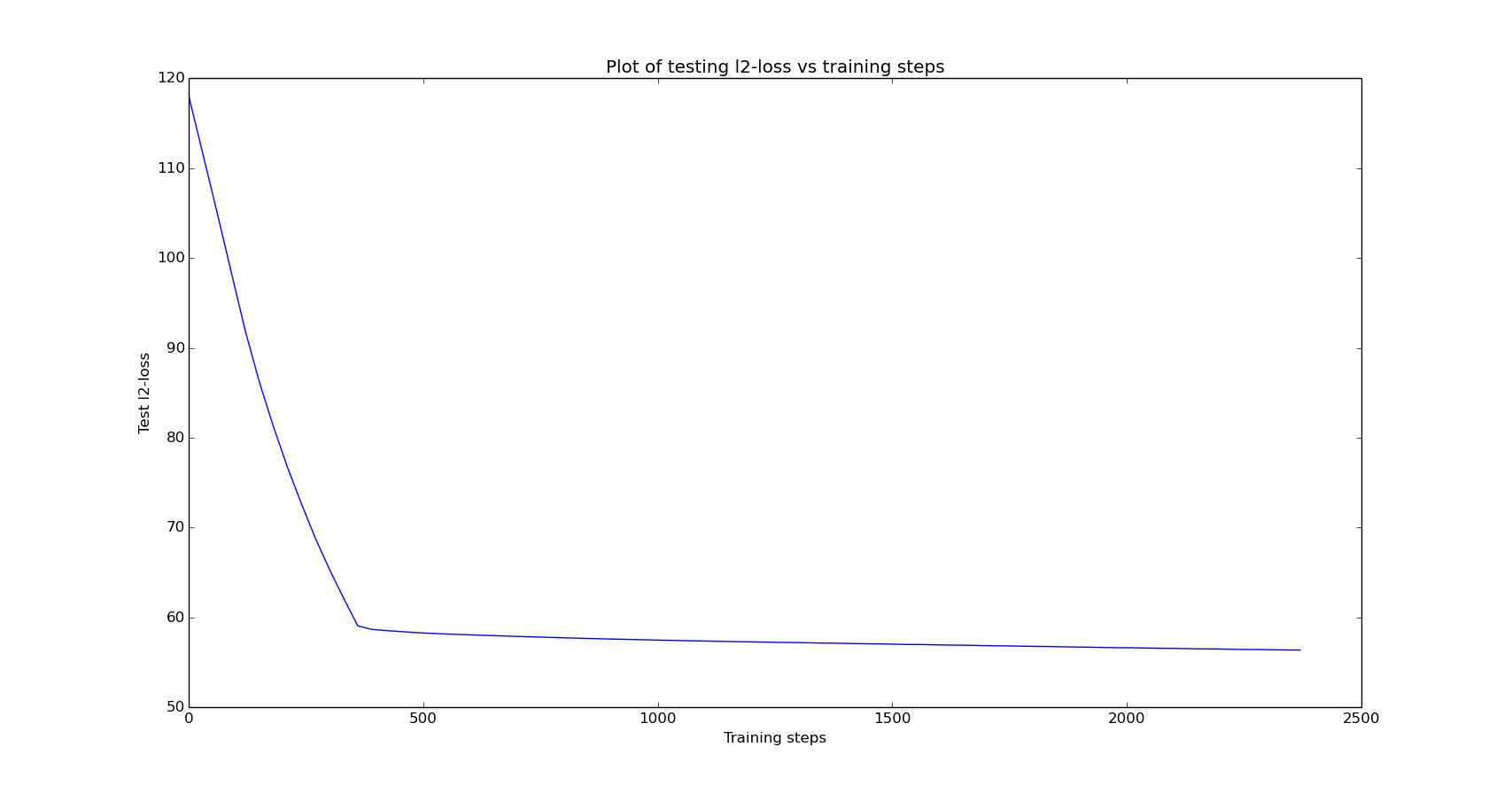}
\caption{Plot of test $L_2$ loss with progress in training}
\end{subfigure}
\caption{$L_2$ loss plots for Subset Selection}
\end{figure}

These results are especially encouraging as the low test losses show the ability of our model to select an optimal subset which incurs the least penalty, i.e. most economical choice with the lowest possible congestion probability.

\section{Related Work}
Many papers previously have tried to tackle these problems intelligently. In \cite{related5}, the paper presents an artificial neural network (ANN) and support vector machine (SVM) approach for locating faults in radial distribution systems. Our approach is novel compared to this paper as we realized that long short-term memories (LSTMs) may be the better way to represent time series voltage and angle data and the hypothesis has been confirmed by our experiments. In \cite{rel6}, a modified PSO (MPSO) mechanism is suggested to deal with the equality and inequality constraints in the ED problems. Our approach is different compared to these existing works as we have used a machine learning and deep learning approach to choose subsets.

\section*{Conclusion}
In conclusion, we created a grid to perform intelligent fault analysis, and subset selection by predicting congestion and selecting a economic choice of subsets having or not having solar generators using deep learning techniques. With the knowledge about the vulnerability of the grid, issues like load shedding, power surges etc. can be handled efficiently. This is expected to be a step towards securing power security in developing countries where clean and reliable power is the requirement for rapid development.

\bibliographystyle{abbrv}
\bibliography{references}

\end{document}